%% file: main.tex
\documentclass{llncs} %

\usepackage{amsmath,amsfonts,amssymb,mathrsfs}
\usepackage{graphicx}
\usepackage{times}
\usepackage[ruled,vlined,linesnumbered]{algorithm2e}
\usepackage{tikz}
\usetikzlibrary{arrows,automata,positioning}
\usepackage{booktabs}
\usepackage{adjustbox}
\usepackage{multirow}
\usepackage{algorithmic}
\usepackage[compatibility=false]{caption}
\usepackage{subcaption}
\usepackage{pifont}
\usepackage{hyperref}

\usepackage{flowchart}
\usetikzlibrary{arrows}
\usepackage{tabularx}

\numberwithin{equation}{section}

\newcommand{\db}[1]{[\kern-0.15em[ #1
 ]\kern-0.15em]}

\makeatletter
\newcommand*{\rom}[1]{\expandafter\@slowromancap\romannumeral #1@}
\makeatother

\newcommand{\myparagraph}[1]{\vspace{1.5ex}\noindent\emph{#1}\hspace{1.5ex}}

\begin{document}

\title{Towards `Verifying' a Water Treatment System\thanks{Corresponding authors: Sun Jun, Shengchao Qin.}}

\author{
 Jingyi Wang\inst{1 3}, Sun Jun\inst{1}, Yifan Jia\inst{1 4}, Shengchao Qin\inst{2 3}, Zhiwu Xu\inst{3}
 \institute{
 $^1$Singapore University of Technology and Design\\
 $^2$School of Computing, Media and the Arts,Teesside University\\ $^3$ College of Computer Science and Software Engineering, Shenzhen University\\ $^4$TUV-SUD Asia Pacific Pte Ltd, Singapore
 }
}

\maketitle
\input{contents/abstract}
\input{contents/introduction}
\input{contents/background}

\input{contents/appr}

\input{contents/results}

\input{contents/related}

\section*{Acknowledgement}
The work was supported in part by Singapore NRF Award No.~NRF2014NCR-NCR001-40, NSFC projects 61772347, 61502308, STFSC project JCYJ20170302153712968.

\clearpage
\bibliographystyle{plain}
\bibliography{reference}
\end{document}

%% file: contents/abstract.tex
\begin{abstract}
Modeling and verifying real-world cyber-physical systems is challenging, which is especially so for complex systems where manually modeling is infeasible. In this work, we report our experience on combining model learning and abstraction refinement to analyze a challenging system, i.e., a real-world Secure Water Treatment system (SWaT). Given a set of safety requirements, the objective is to either show that the system is safe with a high probability (so that a system shutdown is rarely triggered due to safety violation) or not. As the system is too complicated to be manually modeled, we apply latest automatic model learning techniques to construct a set of Markov chains through abstraction and refinement, based on two long system execution logs (one for training and the other for testing). For each probabilistic safety property, we either report it does not hold with a certain level of probabilistic confidence, or report that it holds by showing the evidence in the form of an abstract Markov chain. The Markov chains can subsequently be implemented as runtime monitors in SWaT. 
% This is the first case study of applying model learning techniques to verify a real-world cyber-physical system as far as we know. 
\end{abstract}

%% file: contents/introduction.tex
\section{Introduction}

Cyber-physical systems (CPS) are ever more relevant to people's daily life. Examples include power supply which is controlled by smart grid systems, water supply which is processed from raw water by a water treatment system, and health monitoring systems. CPS often have strict safety and reliability requirements. However, it is often challenging to formally analyze CPS since they exhibit a tight integration of software control and physical processes. Modeling CPS alone is a major obstacle which hinders many system analysis techniques like model checking and model-based testing.

The Secure Water Treatment testbed (SWaT) built at Singapore University of Technology and Design~\cite{swat} is a scale-down version of an industry water treatment plant in Singapore. The testbed is built to facilitate research on cyber security for CPS, which has the potential to be adopted to Singapore's water treatment systems. SWaT consists of a modern six-stage process. The process begins by taking in raw water, adding necessary chemicals to it, filtering it via an Ultrafiltration (UF) system, de-chlorinating it using UV lamps, and then feeding it to a Reverse Osmosis (RO) system. A backwash stage cleans the membranes in UF using the water produced by RO. The cyber portion of SWaT consists of a layered communications network, Programmable Logic Controllers (PLCs), Human Machine Interfaces (HMIs), Supervisory Control and Data Acquisition  (SCADA) workstation, and a Historian. Data from sensors is available to the SCADA system and recorded by the Historian for subsequent analysis. There are 6 PLCs in the system, each of which monitors one stage using a set of sensors embedded in the relevant physical plants and controls the physical plants according to predefined control logics. SWaT has a strict set of safety requirements (e.g., the PH value of the water coming out of SWaT must be within certain specific range). In order to guarantee that the safety requirements are not violated, SWaT is equipped with safety monitoring devices which trigger a pre-defined shutdown sequence. Our objective is thus to show that the probability of a safety violation is low and thus SWaT is reliable enough to provide its service.
%SWaT is highly nontrivial to analyze due to the interaction between the continuous nature of the physical environment and the discrete controller.

One approach to achieve our objective is to develop a model of SWaT and then apply techniques like model checking. Such a model would have a discrete part which models the PLC control logic and a continuous part which models the physical plants (e.g., in the form of differential equations). Such an approach is challenging since SWaT has multiple chemical processes. For example, the whole process is composed of pre-treatment, ultrafiltration and backwash, de-chlorination, reverse osmosis and output of the processed water. The pre-treatment process alone includes chemical dosing, hydrochloric dosing, pre-chlorination and salt dosing. Due to the complexity in chemical reactions, manual modeling is infeasible. Furthermore, even if we are able to model the system using modeling notations like hybrid automata~\cite{henzinger2000theory}, the existing tools/methods~\cite{platzer2010logical,gao2014delta,niggemann2012learning} for analyzing such complicated hybrid models are limited.

%Hybrid automata~\cite{henzinger2000theory} is a popular way to model the interaction of the continuous environment and the discrete controller. The combined specification of discrete and continuous behaviors enables modeling and analysis of dynamic systems like Cyber-Physical systems~\cite{}. However, it also demands that a model is given beforehand, which is challenging for SWaT as explained.

An alternative approach which does not require manual modeling is statistical model checking (SMC)~\cite{YounesThesis,legay2010statistical,clarke2011statistical}. The main idea is to observe sample system executions and apply standard techniques like hypothesis testing to estimate the probability that a given property is satisfied. SMC however is not ideal for two reasons. First, SMC treats the system as a black box and does not provide insight or knowledge of the system on why a given property is satisfied. Second, SMC requires sampling the system many times, whereas starting/restarting real-world CPS like SWaT many times is not viable. 

Recently, there have been multiple proposals on applying model learning techniques to automatically `learn' system models from system executions and then analyze the learned model using techniques like model checking. A variety of learning algorithms have been proposed (e.g.,~\cite{sen2004learning,ron1996power,carrasco1994learning,niggemann2012learning}), some of which require only a few system executions. These approaches offer an alternative way of obtaining models, when having a model of such complex systems is a must. For instance, in~\cite{mao2016learning,chen2012learning,wang2017should,op2}, it is proposed to learn a probabilistic model first and then apply Probabilistic Model Checking (PMC) to calculate the probability of satisfying a property based on the learned model. 

It is however far from trivial to apply model learning directly on SWaT. Existing model learning approaches have only been applied to a few small benchmark systems. It is not clear whether they are applicable or scalable to real-world systems like SWaT. In particular, there are many sensors in SWaT, many of which generate values of type float or double. As a result, the sensor readings induce an `infinite' alphabet which immediately renders many model learning approaches infeasible. In fact, existing model learning approaches have rarely discussed the problem of data abstraction. To the best of our knowledge, the only exception is the LAR method~\cite{op2}, which proposes a method of combining model learning and abstraction/refinement. However, LAR requires many system executions as input, which is infeasible in SWaT.
In this work, we adapt the LAR method so that we require only two long sequences of system execution logs (one for training and the other for testing) as input. 
%The system execution log is a public dataset composed of the values of all the embedded sensors and actuators in the plants of SWaT from a week's normal operation.
We successfully `verified' most of the properties for SWaT this way. For each property, we either report that the property is violated with a certain confidence, or report that the property is satisfied, in which case we output a model in the form of an abstract Markov chain as evidence, which could be further validated by more system runs or expert review. Note that in practice these models could be implemented as runtime monitors in SWaT. %They could also serve as simulative model of the physical environment of SWaT for other analysis tasks like cyber attack detection~\cite{kang2016model}.
%According to the results, we successfully `verified' most of the properties. 
% As far as we know, this is the first real-world application of model learning techniques to analyze CPS systems like SWaT.

The remainders of the paper are organized as follows. Sec.~\ref{back} presents background on SWaT, our objectives as well as some preliminaries. Sec.~\ref{apprs} details our learning approach. We present the results in Sec.~\ref{result} and conclude with related work in Sec.~\ref{related}.

%% file: contents/background.tex
\section{Background}
\label{back}
In this section, we present the target SWaT system and state our motivation and goals.

\subsubsection{System Overview} The system under analysis is the Secure Water Treatment (SWaT) built at the iTrust Center in Singapore University of Technology and Design~\cite{mathur2016swat}. It is a testbed system which scales down but fully realized the functions of a modern water treatment system in cities like Singapore. It enables researchers to better understand the principles of cyber-physical Systems (CPS) and further develop and experiment with smart algorithms to mitigate potential threats and guarantee its safety and reliability.

\begin{figure}[t]
	\centering
	\includegraphics[width=.95\textwidth]{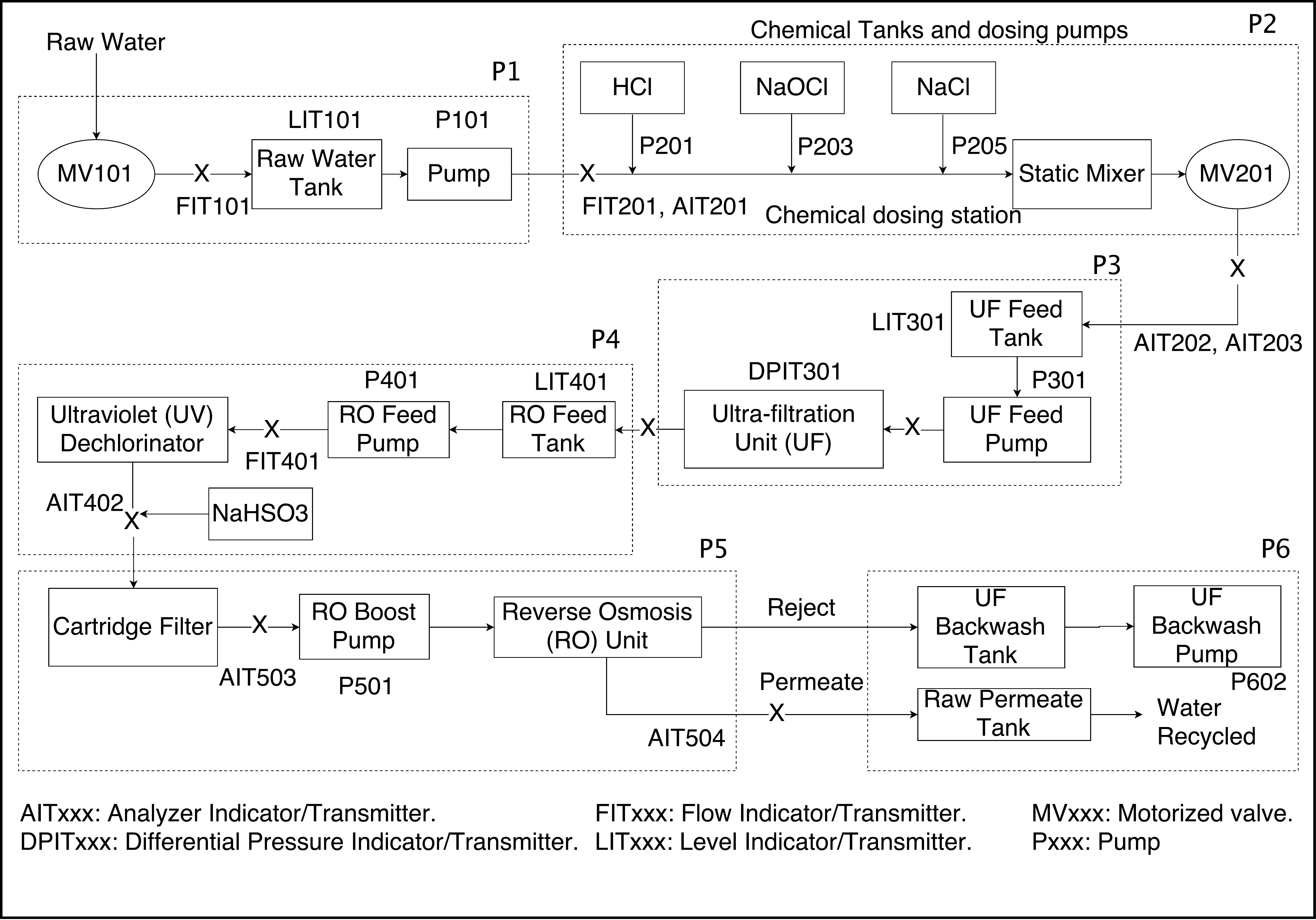}
	\caption{Six stages of water treatment in SWaT~\cite{mathur2016swat}.}
	\label{fig:plants}
\end{figure}

SWaT takes raw water as input and executes a series of treatment and output recycled water eventually. The whole process contains 6 stages as shown in Figure~\ref{fig:plants}. The raw water is taken to the raw water tank (P1) and then pumped to the chemical tanks. After a series of chemical dosing and a static mixer (P2), the water is filtered by an Ultra-filtration (UF) system (P3) and UV lamps (P4). It is then fed to a Reverse Osmosis (RO) system (P5) and a backwash process cleans the membranes in UF using the water produced by RO (P6). For each stage, a set of sensors are employed to monitor the system state. Meanwhile, a set of actuators controlled by the programming logic controller (PLC) are built in to manipulate the state of the physical process. The readings of sensors are collected and sent periodically to the PLC, while the PLC returns a set of actuators values according to the control logics and the current sensor values. For instance, the sensor $LIT101$ is used to monitor the water level of the Raw Water Tank. The PLC reads its value and decides whether to set a new value to the actuators. For example if $LIT101$ is beyond a threshold, the PLC may deactivate the valve $MV101$ to stop adding water into the tank.

SWaT has many built-in safety mechanisms enforced in PLC. Each stage is controlled by local dual PLCs with approximately hundreds of lines of code. In case one PLC fails, the other PLC takes over. The PLC inspects the received and cached sensor values and decides the control strategy to take. Notice that the sensor values are accessible across all PLCs. For example, the PLC of tank 1 may decide whether to start pump $P101$ according to the value of $LIT301$, i.e., the water level of tank 3. In case the controller triggers potential safety violations of the system according to the current values of the sensors, the controller may shut down the system to ensure the safety. The system then needs to wait for further inspection from technicians or experts. Shutting down and restarting SWaT however is highly non-trivial, which takes significant costs in terms of both time and resource, especially in the real-world scenario. Thus, \textit{instead of asking whether a safety violation is possible, the question becomes: how often a system shutdown is triggered due to potential safety violations?} %If we know the system violates the safety property only with very low probability, we may take milder actions to avoid the cost of shutting down the system and vice versa.

In total, SWaT has 25 sensors (for monitoring the status) and 26 actuators (for manipulating the plants). Each sensor is designed to operate in a certain safe range. If a sensor value is out of the range, the system may take actions to adjust the state of the actuators so that the sensor values would go back to normal. Table~\ref{tb:prop} shows all the sensors in the 6 plants, their operation ranges. The sensors has 3 categories distinguished by their prefixes. For instance, $AITxxx$ stands for Analyzer Indicator/Transmitter; $DPITxxx$ stands for Differential Pressure Indicator/Transmitter; $FITxxx$ stands for Flow Indicator Transmitter; $LITxxx$ stands for Level Indicator/Transmitter.

SWaT is also equipped with a historian which records detailed system execution log, including all sensor readings and actuator status. Table~\ref{tb:log} shows a truncated system log with part of sensors. Each row is the sensor readings at a time point and each row is collected every millisecond. Notice that different sensors may have different collection period. The table is filled such that a sensor keeps its old value if no new value is collected, e.g., $AIT202$ in Table~\ref{tb:log}. A dataset of SWaT has been published by the iTrust lab in Singapore University of Technology and Design~\cite{dataset,goh2016dataset}. The dataset contains the execution log of 11 consecutive days (i.e., 7 days of normal operations and another 4 days of the system being under various kind of attacks~\cite{dataset,goh2016dataset}). %The goal the dataset is to assist testing and evaluation for CPS research. For modeling purpose, we only use the data collected from 7 days' normal operations.

\begin{table}[t]
\centering
\caption{Safety properties.}
\label{tb:prop}
\scalebox{1}{
\begin{tabular}{@{}c|ccc@{}}
\toprule
Plant & Sensor & Description                 & Operating range  points                      \\ \midrule
P1    & FIT101         & Flow Transmitter (EMF)         & $2.5-2.6m^3/h$                                         \\
      & LIT101         & Level Transmitter (Ultrasonic) & 500 - 1100$mm$     \\ \midrule
P2    & AIT201         & Analyser (Conductivity)        & 30 - 260$\mu S/cm$ \\
      & AIT202         & Analyser (pH)                  & 6-9             \\
      & AIT203         & Analyser (ORP)                 & 200 - 500mV     \\
      & FIT201         & Flow Transmitter (EMF)         & 2.4 - 2.5$m^3/h$   \\ \midrule
P3    & DPIT301        & DP Transmitter                 & 0.1 - 0.3 Bar   \\
      & FIT301         & Flow Transmitter (EMF)         & 2.2 - 2.4m3/    \\
      & LIT301         & Level Transmitter (Ultrasonic) & 800 - 1000mm    \\ \midrule
P4    & AIT401         & Analyser (Hardness)            & 5-30ppm         \\
      & AIT402         & Analyser                       & 150 - 300mV     \\
      & FIT401         & Flow Transmitter (EMF)         & 1.5 - 2$m^3/h$     \\
      & LIT401         & Level Transmitter (Ultrasonic) & 800 - 1000mm    \\ \midrule
P5    & AIT501         & Analyser (pH)                  & 6-8             \\
      & AIT502         & Analyser (ORP)                 & 100-250mV       \\
      & AIT503         & Analyser (Cond)                & 200- 300$\mu S/cm$   \\
      & AIT504         & Analyser (Cond)                & 5-10$\mu S/cm$      \\
      & FIT501         & Flow Transmitter               & 1-2$m^3/h$         \\
      & FIT502         & Flow Transmitter (Paddlewheel) & 1.1 - 1.3$m^3/h$   \\
      & FIT503         & Flow Transmitter (EMF)         & 0.7 - 0.9$m^3/h$   \\
      & FIT504         & Flow Transmitter (EMF)         & 0.25 - 0.35$m^3/h$ \\
      & PIT501         & Pressure Transmitter           & 2-3 Bar         \\
      & PIT502         & Pressure Transmitter           & 0-0.2 Bar       \\
      & PIT503         & Pressure Transmitter           & 1-2 Bar         \\ \bottomrule
\end{tabular}
}
\end{table}

\subsubsection{Objectives}

%\subsubsection{Safety properties}
%SWaT is safety critical and thus it is designed with many safety requirements and measurements in mind. 

As discussed above, each sensor reading is associated with a safe range, which constitutes a set of safety properties (i.e., reachability). We remark that we focus on safety properties concerning the stationary behavior of the system in this work rather than those properties concerning the system initializing or shutting down phase. In general, a stationary safety property (refer to~\cite{chen2012learning} for details) takes the form $\mathcal{S}_{\leq r}(\varphi)$ (where $r$ is the safety threshold and $\varphi$ is an LTL formula). In our particular setting, the property we are interested in is that \textit{the probability that a sensor is out of range (either too high or too low) in the long term is below a threshold.} %We take the safety properties as input of our learning framework. The property guides learning towards only extracting those information relevant to the given property as explained later. %A property is `verified' if we could successfully learn a model which satisfies the property.
Our objective is to `verify' whether a given set of stationary properties are satisfied or not.

Manual modeling of SWaT is infeasible, with 6 water tanks interacting with each other, plenty of chemical reactions inside the tanks and dozens of valves controlling the flow of water. A group of experts from Singapore's Public Utility Board have attempted to model SWaT manually but failed after months of effort because the system is too complicated. We remark that without a system model, precisely verifying the system is impossible. As discussed above, while statistical model checking (SMC) is another option to provide a statistical measure on the probability that a safety property is satisfied, it is also infeasible in our setting. %One reason is that it may require many system runs to draw a conclusion on whether the property is satisfied while we only have a single system execution log for SWaT. Besides, in case a property is verified, it will not provide any insight on why the property is satisfied.

Thus, in this work, we aim to verify the system by means of model learning. That is, given a safety property, either we would like to show that the property is violated with certain level of confidence or the property is satisfied with certain evidence. Ideally, the evidence is in the form of a small abstract model, at the right level-of-abstraction, which could be easily shown to satisfy the property. 
The advantage of presenting the model as the evidence is that the model could be further validated using additional data or through expert review. Furthermore, the models can serve other purposes. Firstly, the models could be implemented as runtime monitors to detect potential safety violations at runtime. Secondly, we could also prevent future safety violations by predictive analysis based on the model and take early actions. %Moreover, the model is learned by automatically finding a proper level of abstraction in terms of the property in concern. The identified abstraction may also help expert understand the system better.

\begin{table}[t]
\centering
\caption{A concrete system log with the last column being the abstract system log after predicate abstraction with predicate $LIT101>1100$.}
\label{tb:log}
\adjustbox{max width=\textwidth}{
            \begin{tabular}{@{}lllllllll||c@{}}
\toprule
$FIT101$   & $LIT101$   & $MV101$ & $P101$ & $P102$  & $AIT201$   & $AIT202$  & $AIT203$  & $FIT201$  & $LIT101>$1100 \\ \midrule
2.470294 & 261.5804 & 2     & 2    & 1    & 244.3284 & 8.19008 & 306.101 & 2.471278 & 0\\
2.457163 & 261.1879 & 2     & 2    & 1    & 244.3284 & 8.19008 & 306.101 & 2.468587 & 0\\
2.439548 & 260.9131 & 2     & 2    & 1    & 244.3284 & 8.19008 & 306.101 & 2.467305 & 0\\
2.428338 & 260.285  & 2     & 2    & 1    & 244.3284 & 8.19008 & 306.101 & 2.466536 & 0\\
2.424815 & 259.8925 & 2     & 2    & 1    & 244.4245 & 8.19008 & 306.101 & 2.466536 & 0\\
2.425456 & 260.0495 & 2     & 2    & 1    & 244.5847 & 8.19008 & 306.101 & 2.465127 & 0\\
2.472857 & 260.2065 & 2     & 2    & 1    & 244.5847 & 8.19008 & 306.101 & 2.464742 & 0\\ \bottomrule
\end{tabular}
}
\end{table}

%% file: contents/appr.tex
\section{Our approach}\label{apprs}
We surveyed existing model learning algorithms (for the purpose of system verification through model checking) and found most existing model learning approaches~\cite{mao2016learning,chen2012learning,wang2017should} are inapplicable in our setting. The reason is that the real-typed (float or double) variables in SWaT lead to an infinite alphabet. The only method which seems feasible is the recently proposed model learning approach called LAR (short for learning, abstraction and refinement) documented in~\cite{op2}, which allows us to abstract sensor readings in SWaT and automatically learn models at a proper level of abstraction based on a counterexample guided abstraction refinement (CEGAR) framework. However, LAR was designed to take many independent execution logs as input whereas we have only few long system logs of SWaT. We thus adapt LAR to $s$LAR which learns system models from a single long system log instead. %We choose LAR out of a range of learning algorithms mainly because LAR is the only option to automatically learn models at a proper level of abstraction based on a counterexample guided abstraction refinement (CEGAR) stylish framework.
In the following, we briefly explain how $s$LAR works. Interested readers are referred to~\cite{op2} for the detailed explanation of LAR.

\begin{figure}[t]
	\centering
	\includegraphics[width=.95\textwidth]{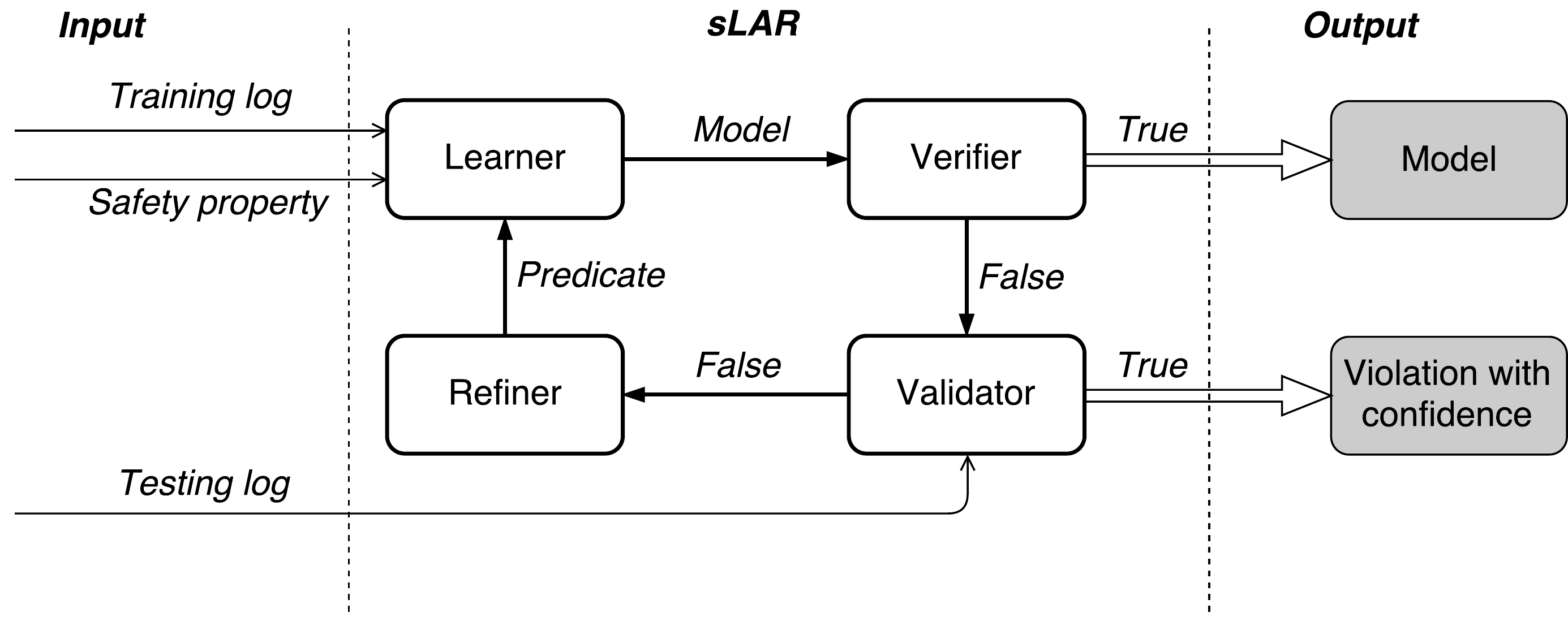}
	\caption{Overall approach.}
	\label{appr}
\end{figure}

Our overall approach is shown in Fig.~\ref{appr}. Given a training log and a safety property, we first construct an abstract log through predicate abstraction and use a learner to learn a model based on the abstract log. Then, the safety property is verified against the learned model. If the verification returns true, we report true and output the learned model as evidence. Otherwise, we test the property using a validator on the testing log. If the validator finds that the property is violated, we report safety violation together with the level of confidence we achieve. Otherwise, we use a refiner to refine the abstraction and start over from the learner. %On termination of our algorithm, we either output a stationary model of the system or report that the safety property is violated with certain confidence. 
Although $s$LAR is based on LAR, our goal of this case study is to verify stationary properties of SWaT and construct a stationary probabilistic model from one single long system log, which is different from LAR. Consequently, the procedures to verify the property and validate the result of the verifier are different. In the following, we present each part of our approach in details.

\subsection{The model}
%We denote the system under analysis (i.e., the SWaT system) by $M$. Assume there are $n$ variables (or sensors\footnote{A sensor is regarded as a variable when modeling}) built in the plants of SWaT, i.e., $V_M=\{V_1,V_2,\cdots,V_n\}$ to monitor the system status. Hereafter, we omit subscript $M$ without ambiguity. Each variable $V_i$ is associated with a valuation domain $D_i$. The domain of a boolean variable $v_b=\{0,1\}$. We write $\Sigma(V)$ (hereafter $\Sigma$ for short) to denote $D_1 \times \cdots \times D_n$, which is the set of all possible variables' valuations of $M$. Note that $\Sigma$ may be infinite if there are real-typed (float or double) variables.

From an abstract point of view, SWaT is a system composed of $n$ variables (including sensors, actuators as well as those variables in the PLC control program) which capture the system status. A system observation $\sigma$ is the valuation of all variables at a time point $t$. A system log $L=\sigma_{t_0}\sigma_{t_1}\cdots\sigma_{t_k}$ is a sequence of system observations collected from time point $t_0$ to $t_k$. Given a system log $L$, we write $L(t)=\sigma_{t}$ to denote the system observation at time $t$ and $L_p(t)$ to denote the system observations before $t$, i.e., from $t_0$ to $t$. In this case study, we use $L$ and $L_t$ to denote the training log and testing log respectively. We also use $T_1$ and $T_2$ to denote their lasting time respectively.

Several machine learning algorithms exist to learn a stationary system model from a single piece of system log~\cite{chen2012learning,ron1996power,wang2017should}. However, applying these algorithms directly is infeasible because of the real-typed (float or double) variables in SWaT, since system observations at different time points are almost always different and thus the input alphabet for the learning algorithms is `infinite'. To overcome this problem, our first step is to abstract the system log through predicate abstraction~\cite{wachter2007probabilistic}. Essentially, a predicate is a Boolean expression over a set of variables. Given a system log and a set of predicates, predicate abstraction turns the concrete variable values to a bit vector where each bit represents whether the corresponding predicate is true or false. For example, given a predicate $LIT101>1100$, the concrete system log on the left of Table~\ref{tb:log} becomes the abstract system log on the right.
%We denote the set of expressions over the set of variables $V$ by $\mathit{Expr}_V$ and the Boolean expressions over $V$ by $\mathit{BExpr}_V$. A predicate $\varphi$ is a Boolean expression. For an expression $e \in \mathit{Expr}_{V}$, we denote its valuation in state $s \in \Sigma$ by $\db{e}_s$. For a predicate $\varphi$, $\db{\varphi}_s\in\{0,1\}$ where 0 stands for \emph{false} and 1 for \emph{true}. We write $s\models\varphi$ iff $\db{\varphi}_s=1$. We denote the set of states satisfying a predicate $\varphi$ by $\db{\varphi}=\{s|s\in\Sigma \land s\models\varphi\}$.
%
%Let $P =\{p_1,\cdots,p_k\} \subseteq \mathit{BExpr}_V$ be a set of predicates over $V$. Given a state $s \in \Sigma$, we define an abstraction function as: $\alpha_P(s)=(\db{p_1}_s,\cdots,\db{p_k}_s)$, which maps an observation $s$ to an abstract observation, i.e., a bit vector with length $k$ where each bit represents the truth value of a predicate in $P$. This effectively reduces an alphabet size bounded by $2^k$, where $k$ is the number of predicates in $P$.
%We write $\Sigma_P$ to denote the set of abstract states with respects to $P$. Given a system log $L = \langle s_1, s_2, \cdots, s_n \rangle$, we construct an abstract system log with respect to predicates $P$ as: $L_P = \langle \alpha_P(s_1), \alpha_P(s_2), \cdots, \alpha_P(s_n) \rangle$.

The models we learn from the log are in the form of discrete-time Markov Chain (DTMC), which is a widely used formalism for modeling stochastic behaviors of complex systems. Given a finite set of states $S$, a probability distribution over $S$ is a function $\mu: S \rightarrow [0,1]$ such that $\sum_{s \in S} \mu(s) = 1$.
Let $\mathit{Distr}(S)$ be the set of all distributions over $S$. Formally,
\begin{definition}
A DTMC $\mathcal{M}$ is a tuple $\langle S,\imath_{init},Pr\rangle$, where $S$ is a countable, nonempty set of states; $\imath_{init}: S \to [0,1]$ is the initial distribution s.t.~$\sum_{s\in S}\imath_{init}(s)=1$; and $Pr: S \rightarrow \mathit{Distr}(S)$ is a transition function such that $Pr(s,s')$ is the probability of transiting from state $s$ to state $s'$. %$L:S\to 2^{AP}$ is a labeling function. $\mathcal{D}$ is \emph{finite} if $S$ is finite.
\end{definition}

% A state of an $s$DTMC represents a finite history of past observations of the system.
% \begin{definition}
% 	An $s$DTMC is a tuple $\mathcal{M}=\langle S, Pr \rangle$, where $S$ is a finite set of states; and $Pr: S \rightarrow \mathit{Distr}(S)$ is a transition function such that $Pr(s,s')$ is the probability of transiting from state $s$ to state $s'$.
% \end{definition}

 We denote a path starting with $s_0$ by $\pi^{s_0} = \langle s_0, s_1, s_2, \cdots, s_n \rangle$, which is a sequence of states in $\mathcal{M}$, where $Pr(s_i,s_{i+1})>0$ for every $0\leq i< n$. 
 % we write $\mathcal{P}(\pi, \mathcal{M}) = Pr(s_0, s_1) \times Pr(s_1, s_2) \times \cdots \times Pr(s_{n-1}, s_n)$ to denote the probability of exhibiting $\pi^{s_0}$ in $\mathcal{M}$. 
 Furthermore, we write $\mathit{Path}^s_{fin}(\mathcal{M})$ to denote the set of finite paths of $\mathcal{M}$ starting with $s$. We say that $s_j\in \pi^{s_0}$ if $s_j$ occurs in $\pi^{s_0}$. In our setting, we use a special form of DTMC, called stationary DTMC (written as $s$DTMC) to model the system behaviors in the long term. Compared to a DTMC,  each state in an $s$DTMC represents a steady state of the system and thus there is no prior initial distribution over the states.

\begin{definition}
	An $s$DTMC is irreducible if for every pair of states $s_i,s_j\in S$ , there exists a path $\pi^{s_i}$ such that $s_j\in \pi^{s_i}$.
\end{definition}

Intuitively, an $s$DTMC is irreducible if there is path between every pair of states. For an irreducible $s$DTMC, there exists a unique stationary probability distribution which describes the average time a Markov chain spends in each state in the long run.

\begin{definition}
	Let $\mu_j$ denote the long run proportion of time that the chain spends in state $s_j$: $\mu_j=\lim_{n\to\infty}\frac{1}{n}\sum_{m=1}^n I\{X_m=s_j|X_0=s_i\}\ with\ probability\ 1.$, for all states $s_i$. If for each $s_j\in S$, $\mu_j$ exists and is independent of the initial state $s_i$, and $\sum_{s_j\in S}\mu_j=1$, then the probability distribution $\mu=(\mu_0,\mu_1,\cdots)$ is called the limiting or stationary or steady-state distribution of the Markov chain.
\end{definition}

In this work, we `learn' a stationary and irreducible $s$DTMC to model the long term behavior of SWaT. By computing the steady-state distribution of the learned $s$DTMC, we can obtain the probability that the system is in the states of interests in the long run.

\subsection{Learning algorithm}
After predicate abstraction, the training log becomes a sequence of bit vectors, which is applicable for learning. We then apply an existing learning algorithm in~\cite{ron1996power} to learn a stationary system model. The initial learned model is in the form of a \emph{Probabilistic Suffix Automata} (PSA) as shown in Figure~\ref{fig:pst}, where a system state in the model is identified by a finite history of previous system observations.
A PSA is an $s$DTMC by definition. Each state in a PSA is labeled by a finite memory of the system.
% Different states could be labeled by labels of different length, which yields varying memory length.
The transition function between the states are defined based on the state labels such that there is a transition $s\times \sigma \to t$ iff $l(t)$ is a suffix of $l(s)\cdot \sigma$, where $l(s)$ is the string label of $s$. A walk on the underlying graph of a PSA will always end in a state labeled by a suffix of the sequence. Given a system log $L_p(t)$ at $t$, a unique state in the PSA can be identified by matching the state label with the suffixes of $L_p(t)$. For example, $\cdots010$ is in state labeled by 0 and if we observe 1 next, the system will go to state labeled by 01.

To learn a PSA, we first construct an intermediate tree representation called \emph{Probabilistic Suffix Tree} (PST), namely $\mathit{tree}(L) = (N, root, E)$ where $N$ is the set of suffixes of $L$; $\mathit{root} = \langle \rangle$; and there is an edge $(\pi_1, \pi_2) \in E$ if and only if $\pi_2 = \langle e \rangle \cdot \pi_1$. Based on different suffixes of the execution, different probabilistic distributions of the next observation will be formed. The central question is how deep should we grow the PST. A deeper tree means that a longer memory is used to predict the distribution of the next observation. The detailed algorithm is shown in Algorithm~\ref{alg:pstlearning}. The tree keeps growing as long as adding children to a current leaf leads to a significant change (measured by K-L divergence) in the probability distribution of next observation (line 5). After we obtain the PST, we transform it into a PSA by taking the leaves as states and define transitions by suffix matching. We briefly introduce the transformation here and readers are referred to Appendix B of~\cite{ron1996power} for more details. For a state $s$ and next symbol $\sigma$, the next state $s'$ must be a suffix of $s \sigma$. However, this is not guaranteed to be a leaf in the learned $T$. Thus, the first step is to extend $T$ to $T'$ such that for every leaf $s$, the longest \emph{prefix} of $s$ is either a leaf or an internal node in $T'$. The transition functions are defined as follows. For each node $s$ in $T\cap T'$ and $\sigma\in\Sigma$, let $Pr'(s,\sigma)=Pr(s,\sigma)$. For each new nodes $s'$ in $T'-T$, let $Pr'(s',\sigma)=Pr(s,\sigma)$, where $s$ is deepest ancestor of $s'$ in $T$. An example PST and its corresponding PSA after transformation is given in Fig.~\ref{fig:pst}. Readers are referred to~\cite{ron1996power} for details.

\input{figs/before.tikz}

\begin{algorithm}[t]
\caption{$Learn\ PST$}\label{alg:pstlearning}
\begin{algorithmic}[1]
%\REQUIRE a sequence $\alpha$ and a parameter $\epsilon>0$
%\ENSURE a PST $T$.
\STATE Initialize $T$ to be a single root node representing $\langle\rangle$; \\
\STATE Let $S = \{\sigma|fre(\sigma, \alpha) > \epsilon\}$ be the candidate suffix set; \\
\WHILE{$S$ is not empty}
\STATE Take any $\pi$ from $S$; Let $\pi'$ be the longest suffix of $\pi$ in $T$; \\
\STATE (B) If $\mathit{fre}(\pi,\alpha) \cdot \sum_{\sigma\in\Sigma} \mathit{Pr}(\pi, \sigma) \cdot \log{\frac{\mathit{Pr}(\pi, \sigma)}{\mathit{Pr}(\pi', \sigma)}}\ge \epsilon$ \\
~~~~add $\pi$ and all its suffixes which are not in $T$ to $T$;
\STATE (C) If $\mathit{fre}(\pi,\alpha) > \epsilon$, add $\langle e \rangle \cdot \pi$ to $S$ for every $e\in\Sigma$ if $\mathit{fre}(\langle e \rangle \cdot \pi, \alpha)>0$;
\ENDWHILE
%\STATE Extend $T$ by adding all missing children of any internal node $\pi$ if $fre(\sigma, \alpha) >0$
%\STATE For each $s\in T$, let $\tilde{\gamma}(\sigma)=\tilde{p}(\sigma|s')$,where $s'$ is the longest suffix of $s$ in $\tilde{T}$
\end{algorithmic}
\end{algorithm}

% \noindent \textbf{Example}
% Assume that the observation so far is $\alpha = \langle \cdots ba \rangle$. Given the tree shown in Figure~\ref{fig:pst}, the next observation is predicted using the probability distribution of its longest suffix in the tree. For instance, the probability of observing $a$ next would be predicted using the probability distribution associated with node $\langle ba \rangle$, which is $Pr_{\langle ba \rangle}(a) = 0.75$. For another example, the predicted probability to generate string $\langle abaa \rangle$ afterwards is computed as: $Pr_{\langle ba \rangle}(a) \cdot Pr_{\langle aa \rangle}(b) \cdot Pr_{\langle b \rangle}(a) \cdot Pr_{\langle ba \rangle}(a)=0.75\cdot 0.75\cdot 0.5\cdot 0.75$.

\subsection{Verification}
Once we learn an $s$DTMC model, we then check whether the learned model satisfies the given safety property. To do so, we first compute the steady-state distribution of the learned model. There are several methods we could use for the calculation including power methods, solving equations or finding eigenvector~\cite{bass2011stochastic}. The steady-state distribution tells the probability that a state occurs in the long run. Once we obtain the steady-state distribution of the learned model, we could then calculate the probability that the system violates the safety property in the long run by summing up the steady-state probability of all unsafe states. Assume $\mu$ is the steady-state distribution, $S_u$ is the set of unsafe states in the learned model and $P_u$ is the probability that the system is in unsafe states in the long run. We calculate the probability of unsafe states as $P_{u}=\sum_{s_i\in S_u}\mu\{s_i\}$. We then check whether the learned model satisfies the safety property by comparing whether $P_u$ is beyond the safety threshold $r$. Take the PSA model in Figure~\ref{fig:pst} as example. The steady-state distribution over states $[1,00,10]$ is $[0.4,0.31,0.29]$. States $1$ is the unsafe state. The steady-state probability that the system is in unsafe states is thus 0.4.

There are two kinds of results. One is that $P_u$ is below the threshold $r$, which means the learned model under current abstraction level satisfies the safety requirement. Then, we draw the conclusion that the system is `safe' and present the learned model as evidence. The soundness of the result can be derived if the learned abstract model simulates the actual underlying model~\cite{probcegar}. However, since the model is obtained through learning from limited data, it is not guaranteed that the result is sound. Nevertheless, the model can be further investigated by validating it against future system logs or reviewed by experts, which we leave to future works. The other result is that the learned model does not satisfy the safety requirement, i.e., the probability of the system being in an unsafe state in the steady-state is larger than the threshold. In such a case, we move to the next step to validate whether the safety violation is introduced by inappropriate abstraction~\cite{op2} or not.

\subsection{Abstraction refinement}
% \myparagraph{Spuriousness Checking}
In case we learn a model which shows that the probability of the system being in unsafe states in long term is beyond the safety threshold, we move on to validate whether the system is indeed unsafe or the violation is spurious due to over-abstraction. For spuriousness checking, we make use of a testing log which is obtained independently and compute the probability of the system being in unsafe states, which is denoted by $P_{u}^t$. The testing log has the same format with the training log. We estimate $P_u^t$ by calculating the frequency that the system is in some unsafe states in the testing log. If $P_u^t$ is larger than the threshold $r$, we report the safety violation together with a confidence by calculating the error bound~\cite{sen2004statistical}. Otherwise, we conclude that the violation is caused by too coarse abstraction and move to the next step to refine the abstraction.

Let $N$ be the total number of states, and $n$ be the number of unsafe states in the testing log. Let $Y=X_1+X_2+\cdots+X_N$, where $X_i$ is a Bernoulli random variable on whether a state is unsafe. The confidence of the safety violation report is then calculated as $\alpha=1-\mathcal{P}\{Y=n|P_u<r\}$. For example, for property $LIT101>1000$, if we observe 1009 times (n) that $LIT101$ is larger than 1000 and the total length of the testing log is 100000 (N), then the estimated $P_u^t$ is $1009/100000=0.01009$.

If we conclude that the current abstraction is too coarse, we continue to refine the abstraction by generating a new predicate following the approach in~\cite{op2}. The predicate is then added to the set of predicates to obtain a new abstract system log based on the new abstraction. The algorithm then starts over to learn a new model based on the new abstract log. Next, we introduce how to generate a new predicate in our setting.

\myparagraph{Finding spurious transitions} A spurious transition in the learned model is a transition whose probability is inflated due to the abstraction. Further, a transition $(s_i,s_{j})$ is spurious if the probability of observing $s_i$ transiting to $s_j$ in the actual system $P_\mathcal{M}(s_i,s_j)$ is actually smaller than $P_{\mathcal{M}_P}(s_i,s_j)$ in the learned model~\cite{op2}. Without the actual system model, we estimate the actual transition probability based on the testing log.
Given the learned model $\mathcal{M}_P$ and the testing log $L_t$, we count the number of times $s_i$ is observed in $L_t$ (denoted by $\#s_i$) and the number of times the transition from $s_i$ to $s_j$ in is observed $L_t$ (denoted by $\#(s_i,s_j)$) using Alg.~\ref{alg:fst}. The actual transition probability $P(s_i,s_j)$ is estimated by $\widehat{P}_\mathcal{M}(s_i,s_j)=\#(s_i,s_j)/\#s_i$. Afterwards, we identify the transitions satisfying $P_{\mathcal{M}_P}(s_i,s_j)-\widehat{P}_\mathcal{M}(s_i,s_j)>0$ as spurious transitions and order them according to the probability deviation.

\begin{algorithm}[t]
\caption{Algorithm $CountST(\mathcal{M}_P,L_t)$}\label{alg:fst}
\begin{algorithmic}[1]
%\REQUIRE a sequence $\alpha$ and a parameter $\epsilon>0$
%\ENSURE a PST $T$.
\STATE Augment each transition $(s_i,s_j)$ in $\mathcal{M}_P$ with a number $\#(s_i,s_j)$ recording how many times we observe such a transition in $L_t$ and initialize them to 0; \\
\STATE Let $t_0$ be the first time that $suffix(L_t(t_0))$ matches a label of a state in $\mathcal{M}_P$ and a time pointer $t=t_0$; \\
\WHILE{$t<T_2$}
\STATE Refer to $\mathcal{M}_P$ for the current state $s_t$; \\
\STATE Take $L_t(t+1)$ from $L_t$ and refer to $\mathcal{M}_P$ to get the next state $s_{t+1}$; \\
\STATE Add $\#(s_t,s_{t+1})$ by 1, add $t$ by 1;
\ENDWHILE
\end{algorithmic}
\end{algorithm}

\myparagraph{Predicate generation} After we obtain a spurious transition $(s_i, s_j)$, our next step is to generate a new predicate to eliminate the spuriousness. The generated predicate is supposed to separate the concrete states of $s_i$ which transit to $s_j$ (positive instances) from those which do not (negative instances). We collect the dataset for classification in a similar way to Alg.~\ref{alg:fst} by iterating the testing log. If $s_i$ is observed, we make a decision on whether it is a positive or negative instance by telling whether its next state is $s_j$. With the labeled dataset, we then apply a supervised classification technique in machine learning, i.e., Support Vector Machines (SVM~\cite{chang2011libsvm,abeel2009java}) to generate a new predicate. Then, we add the predicate for abstraction and start a new round.

\subsection{Overall algorithm}
The overall algorithm is shown as Alg.~\ref{alg:polynomialSVM}. The inputs of the algorithm are a system log $L$ for training, a system log $L_t$ for testing, a property in the form of $\mathcal{S}_{\leq r}(\varphi)$. During each iteration of the loop from line 2 to 16, we start with constructing the abstract trace based on $L$ and a set of predicates $P$. The initial set of predicates for abstraction is the set of predicates in the property. Next, an abstract $s$DTMC $\mathcal{M}_P$ is learned using Algorithm~\ref{alg:pstlearning}. We then verify $\mathcal{M}_P$ against the property. If the property is verified, the system is verified and $\mathcal{M}_P$ is presented as the evidence. Otherwise, we validate the verification result using a testing log $L_t$ at line 9. If the test passes, we report a safety violation together with the confidence. Otherwise, at line 13, we identify the most spurious transition and obtain a new predicate at line 15. After adding the new predicate into $P$, we restart the process from line 2.
If SVM fails to find a classifier for all the spurious transitions, Alg.~\ref{alg:polynomialSVM} terminates and reports the verification is unsuccessful. Otherwise, it either reports true with a supporting model as evidence or a safety violation with confidence.

\begin{algorithm}[t]
    let $P$ be the predicates in $\varphi$\;
    \While {true} {
        construct abstract trace $L_P$ based on training log $L$ and $P$\;
        apply Alg.~\ref{alg:pstlearning} to learn a stationary model $\mathcal{M}_P$ based on $L_P$\;
        check $\mathcal{M}_P$ against $\varphi$\;
        \If {$\mathcal{M}_P \models \varphi$} {
            report $\varphi$ is verified, the model $\mathcal{M}_P$\;
            \Return\;
        }
        use the testing log $L_t$ to validate the property violation\;
        \If {validated} {
            report $\varphi$ is violated with confidence\;
            \Return\;
        }
        identify the most spurious transitions $\langle s, s'\rangle$ in $\mathcal{M}_P$\;
        collect labeled dataset $D^+(s, \mathcal{M}_P, L_t)$ and $D^-(s, \mathcal{M}_P, L_t)$\;
        apply SVM to identify a predicate $p$ separating the two sets\;
        add $p$ into $P$\;
    }
\caption{Algorithm $s$LAR($L$,$L_t$, $\mathcal{S}_{\leq r}(\varphi)$)}
\label{alg:polynomialSVM}
\end{algorithm}

%% file: figs/before.tikz
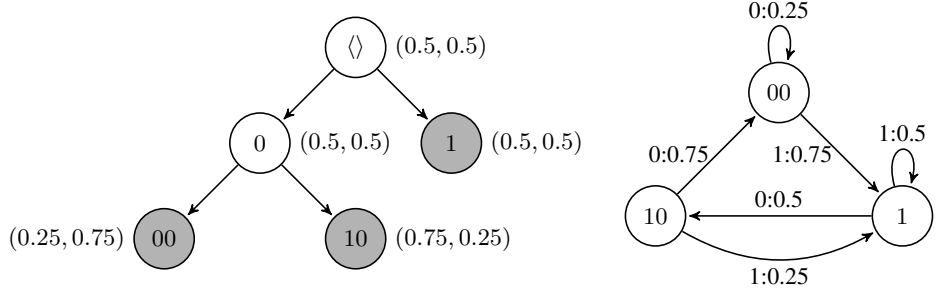
\begin{figure}[t]
  \begin{minipage}[c]{0.4\textwidth}
    \footnotesize{\begin{tikzpicture}[->,>=stealth',shorten >=1pt,auto,node distance=1.8cm,
                      semithick]
    \tikzset{state/.style={circle,draw,minimum size=0.8cm},inner sep=0.1cm}

    \node[state, label=right: {$(0.5,0.5)$}]       (A)                    {$\langle \rangle$};
    \node[state, label=right: {$(0.5,0.5)$}]       (B) [below left of=A]  {$0$};
    \node[state, fill=black!30,label=right: {$(0.5,0.5)$}]       (C) [below right of=A] {$1$};
    \node[state, fill=black!30,label=left: {$(0.25,0.75)$}]     (D) [below left of=B]  {$00$};
    \node[state, fill=black!30,label=right: {$(0.75,0.25)$}]     (E) [below right of=B] {$10$};

    \path (A) edge              node [left] {}  (B)
              edge              node  {}      (C)
          (B) edge              node [left] {}  (D)
              edge              node [right] {}  (E);
  \end{tikzpicture}}
  % \caption{Probabilistic suffix tree.}
  \end{minipage}\hfill
  \begin{minipage}[c]{0.32\textwidth}
  \footnotesize{\begin{tikzpicture}[->,>=stealth',shorten >=1pt,auto,node distance=1.5cm,
semithick]
    \tikzset{state/.style={circle,draw,minimum size=0.8cm},inner sep=0.1cm}
    \node[state]       (A)                    {$00$};
    \node[state]       (B) [below left =1.5cm of A]  {$10$};
    \node[state]       (C) [below right = 1.5cm of A] {$1$};

    \path (A) edge      [loop above]        node  {0:0.25}  (A)
              edge              node [left] {1:0.75}(C)
          (B) edge              node [left] {0:0.75}  (A)
              edge      [bend right]        node [below] {1:0.25}  (C)
          (C) edge              node [above] {0:0.5}  (B)
              edge      [loop above]        node [above] {1:0.5}  (C);

    \end{tikzpicture}}
    % \caption{Transformed DTMC.}
  \end{minipage}
  \caption{An example stationary model. The left is the PST representation, where each state is associated with a label and a distribution of the next observation. The right is the corresponding PSA model where leaves are taken as states.}
  \label{fig:pst}
\end{figure}

%% file: contents/results.tex
\section{Case study results}\label{result}

In the following, we present our findings on applying the method documented in Section~\ref{apprs} to SWaT. Given the 11 day system log~\cite{goh2016dataset}, we take the 7 day log under normal system execution and further split it into two parts for training (4 days) and testing (3 days) respectively. The main reason we split them into training and testing log is to avoid over-fitting problem without the testing data. Note that the historian makes one record every second. The training log and testing log contains 288000 and 208800 system observations respectively. The properties we verified are whether the steady-state probability that a sensor runs out of its operating range is beyond or below a threshold. Let $P_{train}$, $P_{learn}$ and $P_{test}$ be the probability that a sensor is out of operating range in the training log, learned models and the testing log respectively. In our study, we set the threshold $r$ in each property as 20 percent larger than the probability observed in the actual system for a long time, during which the system functioned reliably. The idea is to check whether we can establish some underlying evidence to show that the system would satisfy the property indeed.

The experiment results of all sensors are summarized in Table~\ref{tb:exp}. The detailed implementation and models are available in~\cite{ziqian}. The first column is the plant number. 
Column 2 and 3 are the sensors and their properties to verify which are decided by their operating ranges. The following 4 columns show the probability that a sensor value is out of operating range in the training log, the safety threshold, the probability in the learned model and the probability in the testing log respectively. Column `result' is the verification result of the given safety properties. `SUC' means the property is successfully verified. `FAL' means the property is not verified. `VIO' means the property is violated. Column `model size' is the number of states in the learned model. Column $\epsilon$ is the parameter we use in the learning parameter. The last column is the running time.

\myparagraph{Summary of results} In total, we managed to evaluate 47 safety properties of 24 sensors. Notice that the sensor from P6 is missing in the dataset. Among them, 19 properties are never observed to be violated in the training log. We thus could not learn any models regarding these properties and conclude that the system is safe from the limited data we learn from. This is reasonable as according to the dataset, the probability violating the property is 0. For the rest 28 properties, we successfully verified 24 properties together with a learned abstract Markov chain each and reported 4 properties as safety violation with a confidence. 

We have the following observations from the results. For those properties we successfully verified, we managed to learn stationary abstract Markov chains which closely approximate the steady-state probability of safety violation (evaluated based on the probability computed based on the testing log). It means that in these cases, $s$LAR is able to learn a model that is precise enough to capture how the sensor values change. Examples are $FIT101>2.6$, $LIT301>1000$, $LIT301<800$ and $LIT401>1000$. Besides, it can be observed that the learned abstract models are reasonably small, i.e., usually with less than 100 states and many with only a few states. This is welcomed since a smaller model is easier to comprehend and thus more meaningful for expert review or to be used as a runtime monitor. An underlying reason (why a small model is able to explain why a property is satisfied) is perhaps the system is built such that the system modifies its behavior way before a safety violation is possible. Besides, we identify two groups of states which are of special interest. One of them are $FIT401<1.5$, $FIT502<1.1$, $FIT503<0.7$ and $FIT504<0.25$. The 4 properties have the same probability 0.0117 of safety violation in the training log and 0 in the testing log. We learn the same models for all of them and $P_{learn}$ equals 0 which is the same as the testing log. We could observe that these sensors have tight connections with each other. Moreover, these sensors are good examples that our learned models generalize from the training data and are able to capture the long run behaviors of the system with $P_{learn}$ equals $P_{test}$, which is 0. The same goes for the other group of properties, i.e., $FIT501<1$, $PIT501<20$ and $PIT503<10$.   

For those properties we reported as safety violations, i.e., $AIT401>100$, $PIT501>30$, $PIT502>0.2$ and $PIT503>20$, a closer look reveals that these sensors all have high probability of violation (either 0.7156 or 0.989) in the training log. Our learned models report that the probability of violation in the long term is 1, which equals the probability in the testing log in all cases. This shows that our learned models are precise even though the properties are not actually satisfied.  

% we fail because we could not refine the abstraction (by identifying new predicates). Furthermore, these are often properties which have relatively small probabilities. Our investigation shows that the sensors that we failed to verify have tight connections with each other and have the same trend in the data. For example,   One possible reason for failing to identify the necessary predicate for abstraction refinement is that the actual predicate may relate these sensors in a complexity way which is beyond the learning capability of SVM (which learns linear inequalities). For $AIT203>500$, we failed to verify it because the iteration for refinement exceeds the preset limit 20. However, we observe that the learned models are getting more and more accurate. For the property we reported safety violation, i.e., $FIT502>1.3$, the sensor values change rapidly. Besides, there is significant difference between the probability in the training and testing log, which might indicate a sensor malfunctioning or a system design problem. 
% TODO: can we confirm whether this is a real problem? (for instance, a sensor malfunctioning or a system design problem?)

\myparagraph{Discussions} 1) We give a 20\% margin for the safety threshold in the above experiments. In practice, the actual safety threshold could be derived from the system reliability requirement. In our experiments, we observe that we could increase the threshold to obtain a more abstract model and decrease the threshold to obtain a more detailed model. For instance, we would be more likely to verify a property with a loose threshold. 2) The parameter $\epsilon$ in Algorithm~\ref{alg:pstlearning} effectively controls the size of learned model. A small $\epsilon$ used in the model learning algorithm leads to a learned model with more states by growing a deeper tree. However, it is sometimes non-trivial to select a good $\epsilon$~\cite{wang2017should}. In our experiment, we use 0.01 as the basic parameter. If we can not learn a model (the tree does not grow), we may choose a more strict $\epsilon$. Examples are $LIT401>1000$ and $AIT504>10$. This suggests one way to improve existing model learning algorithms. 3) Each sensor has a different collection period and most of them are changing very slowly, thus the data is not all meaningful to us and we only take a data point from the dataset every minute to reduce the learning cost. 4) One possible reason for the safety violation cases is that the system has not exhibited stationary behaviors within 7 days as the probability of safety violations is 1 in the testing data for all these cases.

\myparagraph{Limitation and future work} Model learning will correctly learn an underlying model in the limit~\cite{mao2012learning,ron1996power}. However, since our models are learned from a limited amount of data from a practical point of view, they are not guaranteed to converge to the actual underlying models. One of our future work is how to further validate and update the learned models from more system logs. In general, it is a challenging and interesting direction to derive a confidence for the learned model (as a machine learning problem) or the verification results based on the learned models (as a model checking problem) given specific training data. Or alternatively, how can we derive a requirement on the training data to achieve a certain confidence. Some preliminary results on the number of samples required to achieve an error bound are discussed in~\cite{jegourel2017sequential}.  

\begin{table}[t]
\centering
\caption{Experiment results.}
\label{tb:exp}
\scalebox{1}{
\begin{tabular}{@{}c|cccccccccc@{}}
\toprule
Plant & Sensor  & Property         & $P_{train}$ & $r$ & $P_{learn}$ & $P_{test}$ & Result & Model Size & $\epsilon$ & Time \\ \midrule
P1    & FIT101  & \textgreater2.6  & 0.2371  & 0.2845 & 0.2371    & 0.233     & SUC    & 26         & 0.01 & 300  \\
      &         & \textless2.5     & 0.5092  & 0.611 & 0.5092    & 0.5245    & SUC    & 31         & 0.01              & 298  \\
      & LIT101  & \textgreater800  & 0.1279  & 0.1535 & 0.1271    & 0.1141    & SUC    & 130        & 0.01              & 4    \\
      &         & \textless500     & 0.1485  & 0.1782 & 0.147    & 0.0977    & SUC    & 54         & 0.01          & 2    \\ \midrule
P2    & AIT201  & \textgreater260  & 0.6044  & 0.7253 & 0.647       & 1         & SUC    & 2          & 0.01              & 31   \\
      &         & \textless250     & 0       & -- & --        & --        & --     & --         & --              & --   \\
      & AIT202  & \textgreater9    & 0       & -- & --        & --        & --     & --         & --               & --   \\
      &         & \textless6       & 0       & -- & --        & --        & --     & --         & --               & --   \\
      & AIT203  & \textgreater500  & 0.0362  & 0.043 & 0.0363    & 0         & SUC    & 2         & 0.01             & 27 \\
      &         & \textless420     & 0.7654  & 0.9185 & 0.7654       & 1         & SUC    & 2          & 0.01              & 32   \\
      & FIT201  & \textgreater2.5  & 0       & -- & --        & --        & --     & --         & --              & --   \\
      &         & \textless2.4     & 0.2577  & 0.3092 & 0.2567    & 0.2529    & SUC    & 59         & 0.01              & 4    \\ \midrule
P3    & DPIT301 & \textgreater30   & 0       & -- & --        & --        & --     & --         & --               & --   \\
      &         & \textless10      & 0.2006  & 0.2407 & 0.1991    & 0.1799    & SUC    & 119         & 0.01              & 4   \\
      & FIT301  & \textgreater2.4  & 0       & -- & --        & --        & --     & --         & --               & --   \\
      &         & \textless2.2     & 0.2217  & 0.266 & 0.2209    & 0.1756    & SUC    & 42         & 0.01              & 4   \\
      & LIT301  & \textgreater1000 & 0.134   & 0.1608 & 0.135      & 0.1299    & SUC    & 60         & 0.01              & 4    \\
      &         & \textless800     & 0.0877  & 0.1052 & 0.0876    & 0.0609    & SUC    & 69         & 0.01              & 2    \\ \midrule
P4    & AIT401  & \textgreater100  & 0.7156  & 0.8587 & 1       & 1         & VIO    & 2          & 0.002             & 35   \\
      &         & \textless5       & 0.2844  & 0.3413 & 0       & 1         & SUC    & 2          & 0.01              & 33   \\
      & AIT402  & \textgreater250  & 0       & -- & --        & --        & --     & --         & --               & --   \\
      &         & \textless150     & 0       & -- & --        & --        & --     & --         & --               & --   \\
      & FIT401  & \textgreater2    & 0       & -- & --        & --        & --     & --         & --               & --   \\
      &         & \textless1.5     & 0.0117  & 0.014 & 0       & 0         & SUC    & 2          & 0.01              & 37   \\
      & LIT401  & \textgreater1000 & 0.0035  & 0.0042 & 0.0037    & 0.0034    & SUC    & 208        & 0.002             & 455 \\
      &         & \textless800     & 0.1227  & 0.1472 & 0.123       & 0.079     & SUC    & 70         & 0.01              & 2    \\ \midrule
P5    & AIT501  & \textgreater8    & 0       & -- & --        & --        & --     & --         & --               & --   \\
      &         & \textless6       & 0       & -- & --        & --        & --     & --         & --              & --   \\
      & AIT502  & \textgreater250  & 0       & -- & --        & --        & --     & --         & --              & --   \\
      &         & \textless100     & 0       & -- & --        & --        & --     & --         & --               & --   \\
      & AIT503  & 300              & 0       & -- & --        & --        & --     & --         & --              & --   \\
      &         & \textless200     & 0       & -- & --        & --        & --     & --         & --              & --   \\
      & AIT504  & \textgreater10   & 0.9983  & 1 & 0.9983       & 1         & SUC    & 2          & 0.001             & 37   \\
      &         & \textless5       & 0       & -- & --        & --        & --     & --         & --               & --   \\
      & FIT501  & \textgreater2    & 0       & -- & --        & --        & --     & --         & --               & --   \\
      &         & \textless1       & 0.011   & 0.0132 & 0    & 0         & SUC    & 3          & 0.01              & 38   \\
      & FIT502  & \textgreater1.3  & 0.0356  & 0.0427 & 0.0361    & 0.3241    & SUC    & 9          & 0.01              & 15   \\
      &         & \textless1.1     & 0.0117  & 0.014 & 0       & 0         & SUC    & 2          & 0.01             & 38   \\
      & FIT503  & \textgreater0.9  & 0       & -- & --        & --        & --     & --         & --               & --   \\
      &         & \textless0.7     & 0.0117  & 0.014 & 0       & 0         & SUC    & 2          & 0.01              & 38   \\
      & FIT504  & \textgreater0.35 & 0       & -- & --        & --        & --     & --         & --              & --   \\
      &         & \textless0.25    & 0.0117  & 0.014 & 0       & 0         & SUC    & 2          & 0.01             & 38   \\
      & PIT501  & \textgreater30   & 0.989   & 1 & 1    & 1         & VIO    & 3          & 0.01             & 38   \\
      &         & \textless20      & 0.011   & 0.0132 & 0    & 0         & SUC    & 3          & 0.01             & 38  \\
      & PIT502  & \textgreater0.2  & 0.989  & 1 & 1    & 1         & VIO    & 3          & 0.01              & 37   \\
      & PIT503  & \textgreater20   & 0.989   & 1 & 1    & 1         & VIO    & 3          & 0.01            & 37   \\
      &         & \textless10      & 0.011   & 0.0132 & 0    & 0         & SUC    & 3          & 0.01             & 38   \\ \bottomrule
\end{tabular}
}
\end{table}

%% file: contents/related.tex
\section{Conclusion and related work}\label{related}

In this work, we conducted a case study to automatically model and verify a real-world water treatment system testbed. Given a set of safety properties of the system, we combine model learning and abstraction refinement to learn a model which 1) describes how the system would evolve in the long run and 2) verifies or falsifies the properties. The learned models could also be used for further investigation or other system analysis tasks such as probabilistic model checking, simulation or runtime monitoring.

This work is inspired by the recent trend on adopting machine learning to automatically learn models for model checking. Various kinds of model learning algorithms have been investigated including continuous-time Markov Chain~\cite{sen2004learning}, DTMC~\cite{mao2016learning,chen2012learning,wang2017should,wang2017improving,wang2018learning} and Markov Decision Process~\cite{mao2012learning,brazdil2014verification}. In particular, this case study is closely related to the learning approach called LAR documented in~\cite{op2}, which combines model learning and abstraction refinement to automatically find a proper level of abstraction to treat the problem of real-typed variables. Our algorithm is a variant of LAR, which adapts it to the setting of stationary probabilistic models~\cite{chen2012learning}.

This case study aims to formally and automatically analyze a real-world CPS by modeling and verifying the physical environment probabilistically. There are several related approaches for this goal. One popular way is to model the CPS as hybrid automata~\cite{henzinger2000theory}. In~\cite{platzer2010logical}, a theorem prover for hybrid systems is developed. \emph{dReach} is another tool to verify the $\delta$-complete reachability analysis of hybrid system~\cite{gao2014delta}. Nevertheless, they both require users to manually write a hybrid model using differential dynamic logic, which is highly non-trivial. In~\cite{niggemann2012learning}, the authors propose to learn hybrid models from a sample of observations. In addition, \emph{HyChecker} borrows the idea of concolic testing to hybrid system based on a probabilistic abstraction of the hybrid model and achieves faster detection of counterexamples~\cite{kong2016towards}. $s$LAR is different as it is fully automatic without relying on a user-provided model. SMC is another line of work which does not require a model beforehand~\cite{clarke2011statistical}. However, it requires sampling the system many times. This is unrealistic for our setting since shutting down and restarting SWaT yield significant cost. Besides, SMC does not provide insight on how the system works but only provides the verification result. Our learned models however can be used for other system analysis tasks.

Several case studies are related to our case study in some way. In~\cite{lin2009towards}, the authors applied integrated simulation of the physical part and the cyber part to an intelligent water distribution system. In~\cite{fiteruau2016combining}, the authors use model learning to infer models of different software components for TCP implementation and apply model checking to explore the interaction of different components. In~\cite{kim2013parallel}, a case study on self-driving car is conducted for the analysis of parallel scheduling for CPS. In~\cite{misra2013learning}, automata learning is applied in different levels of a smart grid system to improve the power management.
As far as we know, our work is the first on applying probabilistic model learning for verifying a real-world CPS probabilistically.